\documentclass[11pt]{article}

\usepackage{amsmath,amssymb,graphicx,color,amsthm}
\usepackage[linesnumbered,ruled]{algorithm2e}
\usepackage{tikz}

\usetikzlibrary{shapes.geometric, arrows}
\usepackage{algorithmic}

\usepackage[english]{babel}
\usepackage[utf8x]{inputenc}
\usepackage{subcaption,afterpage} 
\usepackage{mathtools}
\usepackage{amsfonts}
\usepackage[colorinlistoftodos]{todonotes}
\usepackage{enumerate}
\usepackage[round]{natbib}
\usepackage[T1]{fontenc}
\usepackage{hyperref}
\usepackage{natbib}
\usepackage{siunitx}
\newcommand{\excise}[1]{}

\newcommand\RR{\mathbb{R}}
\newcommand\PP{\mathbb{P}}

\newtheorem{theorem}{Theorem}

\newcommand{\RNum}[1]{\uppercase\expandafter{\romannumeral #1\relax}}

\begin{document}
	
	\title{\mbox{}\\[-11ex]Classification via local manifold approximation}
	\vspace{-8ex}
	\author{\\[1ex]Didong Li$^1$ and David B Dunson$^{1,2}$ \\ 
		{\em Department of Mathematics$^1$ and Statistical Science$^2$, Duke University}}
	\date{\vspace{-5ex}}
	
	\maketitle
	Classifiers label data as belonging to one of a set of groups based on input features.  It is challenging to obtain accurate classification performance when the feature distributions in the different classes are complex, with nonlinear, overlapping and intersecting supports.  This is particularly true when training data are limited.  To address this problem, this article proposes a new type of classifier based on obtaining a local approximation to the support of the data within each class in a neighborhood of the feature to be classified, and assigning the feature to the class having the closest support.  This general algorithm is referred to as LOcal Manifold Approximation (LOMA) classification.  As a simple and theoretically supported special case having excellent performance in a broad variety of examples, we use spheres for local approximation, obtaining a SPherical Approximation (SPA) classifier.  We illustrate substantial gains for SPA over competitors on a variety of challenging simulated and real data examples.

	\section{Introduction}
	
	Classification is one of the canonical problems in statistics and machine learning.  In the typical setting, the focus is on learning a classifier 
	$F: \mathcal{X} \to \mathcal{Y} = \{1,\ldots,L\}$ mapping from features $x \in \mathcal{X}$ to class labels $y \in \mathcal{Y}$.  Classifier learning relies on training data $(x_i,y_i)$, $i=1,\ldots,n$, in which both the features $x_i$ and labels $y_i$ are known.  This article focuses on developing methods for difficult classification problems in which the dimension $D$ of the features is not small, the amount of training data $n$ may be limited, and the distributions of the features within the different classes are non-linear, intersecting and `entangled’ with each other.
	
	To clarify the conceptual challenges, we introduce some notation and two motivating toy problems; refer to Figure 1.  Let $\rho_y$ denote the conditional density of the features $x$ given class label $y$, and let $\mathcal{X}_y^{(\epsilon)} = \{ x: \rho_y(x) > \epsilon \}$ denote a region of feature space having high density for data in class $y$.  Classification is a relatively easy problem when the regions $\mathcal{X}_y^{(\epsilon)}$ can be separated with hyperplanes, as in the example in Figure 1a.  In this case, standard classifiers such as logistic regression (\cite{logistic}) and support vector machines (\cite{cortes1995support}) tend to have excellent performance.  However, it is increasingly common in modern applications to be faced with cases more similar to Figure 1b.  In this example, the supports $\mathcal{X}_1^{(\epsilon)}$ and $\mathcal{X}_2^{(\epsilon)}$ have minimal overlap, so that accurate classification performance is conceptually possible. However, most existing algorithms are incapable of learning the classification boundaries accurately, particularly when training data are limited.
	
	\begin{figure}
		\centering
		\begin{subfigure}[t]{0.45\textwidth}
			\centering
			\includegraphics[width=\linewidth]{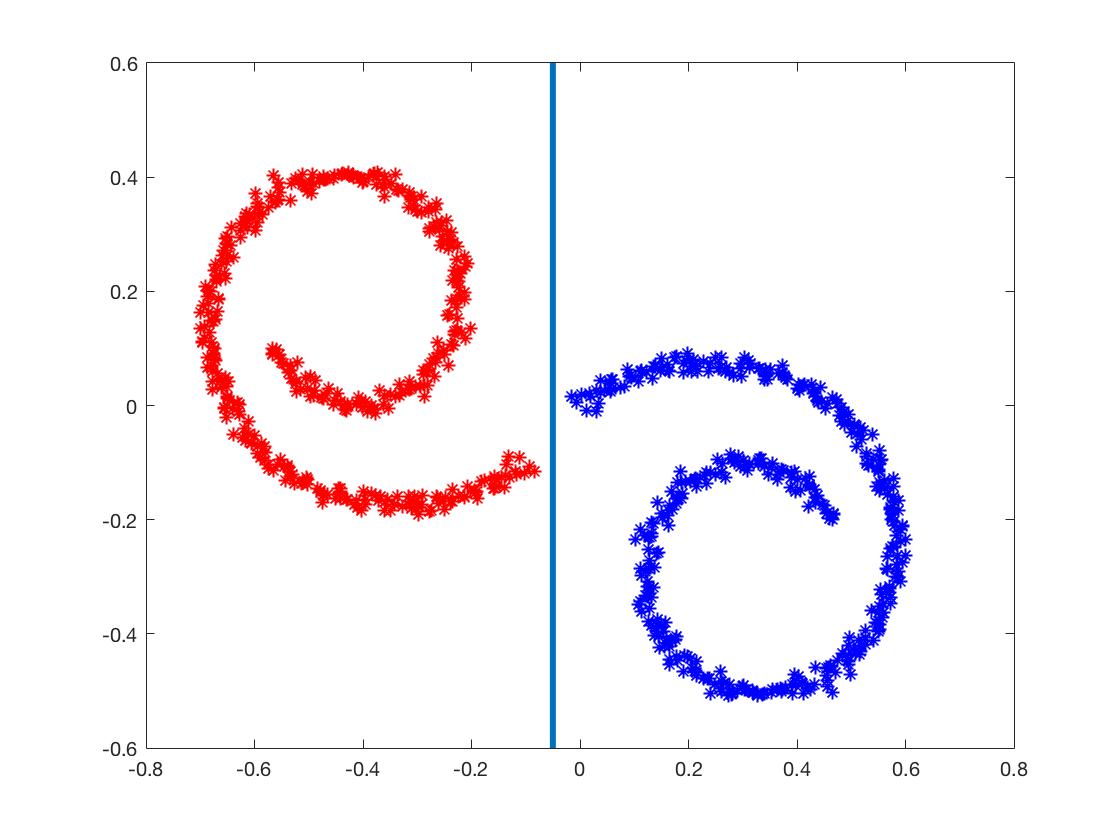} 
			\caption{Linearly separable classes} 
			\label{sep}
		\end{subfigure}
		\hfill
		\begin{subfigure}[t]{0.45\textwidth}
			\centering
			\includegraphics[width=\linewidth]{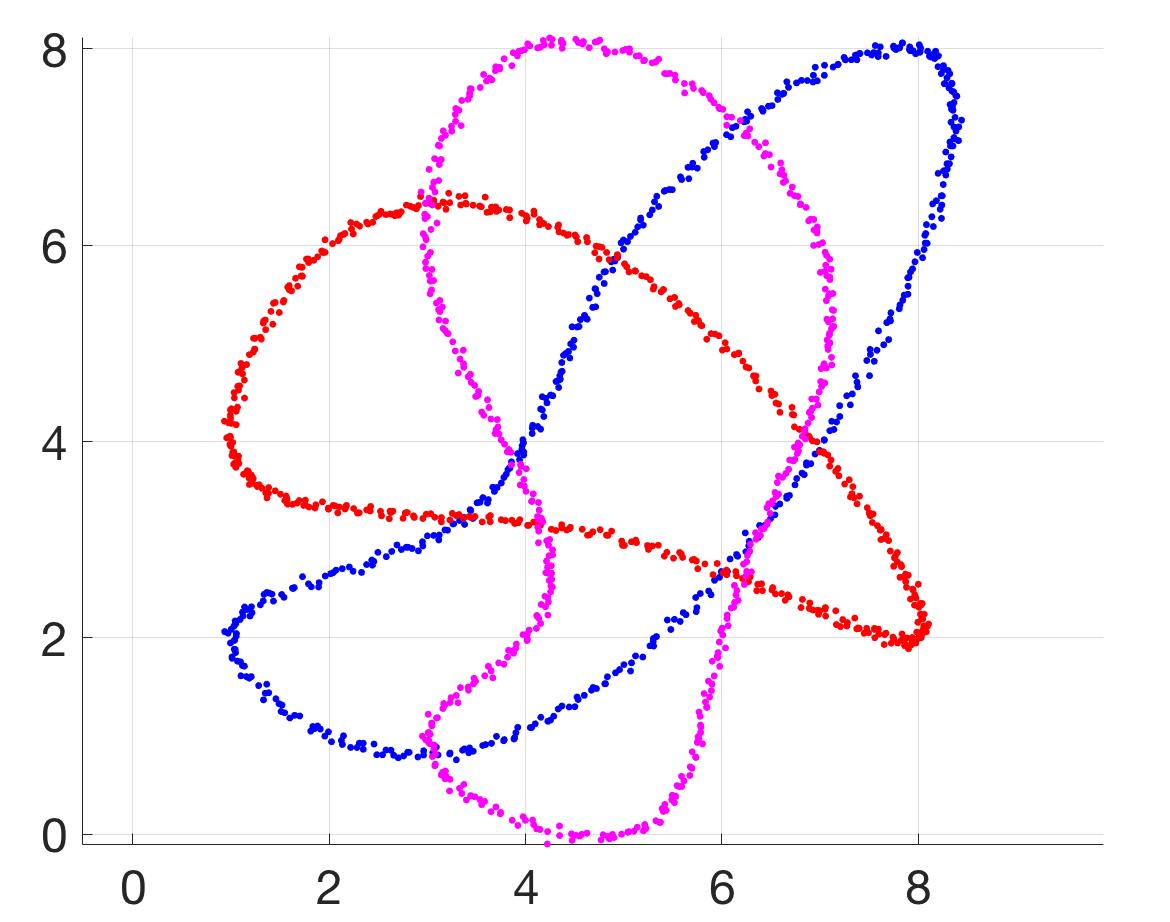} 
			\caption{Linearly inseparable classes} 
			\label{insep}
		\end{subfigure}

		\caption{Two motivating examples}
	\end{figure}
	
			
	
	In recent years, there has been abundant focus on classifiers based on deep learning and multilayer neural networks (refer, for example to \cite{schmidhuber2015deep} and \cite{ciresan2011cnn}). Such approaches have particularly outstanding performance in imaging and in other structured data settings. Key advantages of neural networks include their amazing flexibility, very high learning capacity and classification accuracy under careful tuning and architecture design utilizing large training data sets.  However, this great flexibility leads to challenges in cases with limited training data.  Limited data makes estimation of the very many neural network parameters problematic even utilizing the rich variety of dimensionality reduction tools that have been developed in recent years.  Potentially the architecture can be simplified but then classification accuracy can suffer, either due to a decrease in representation capacity or to over-fitting when the choice of architecture is driven by training data performance.
	
	We propose a simple class of algorithms to overcome these problems in limited data settings.  In particular, suppose we wish to estimate $F(x)$ for some arbitrary feature vector $x=(x_1,\ldots,x_D)^\top$. We first locally approximate the denoised support of the feature density $\rho_y$ within class $y$, for each $y \in \mathcal{Y}$, and then set $F(x)$ equal to the class $y$ having the minimal Euclidean distance between $x$ and the denoised support estimate.  This approach is illustrated in Figure 2 for the example in Figure 1b.  We refer to this general approach as LOcal Manifold Approximation (LOMA) classification.  
	
		\begin{figure}
		\begin{center}
			\includegraphics[height =180pt, width=200 pt]{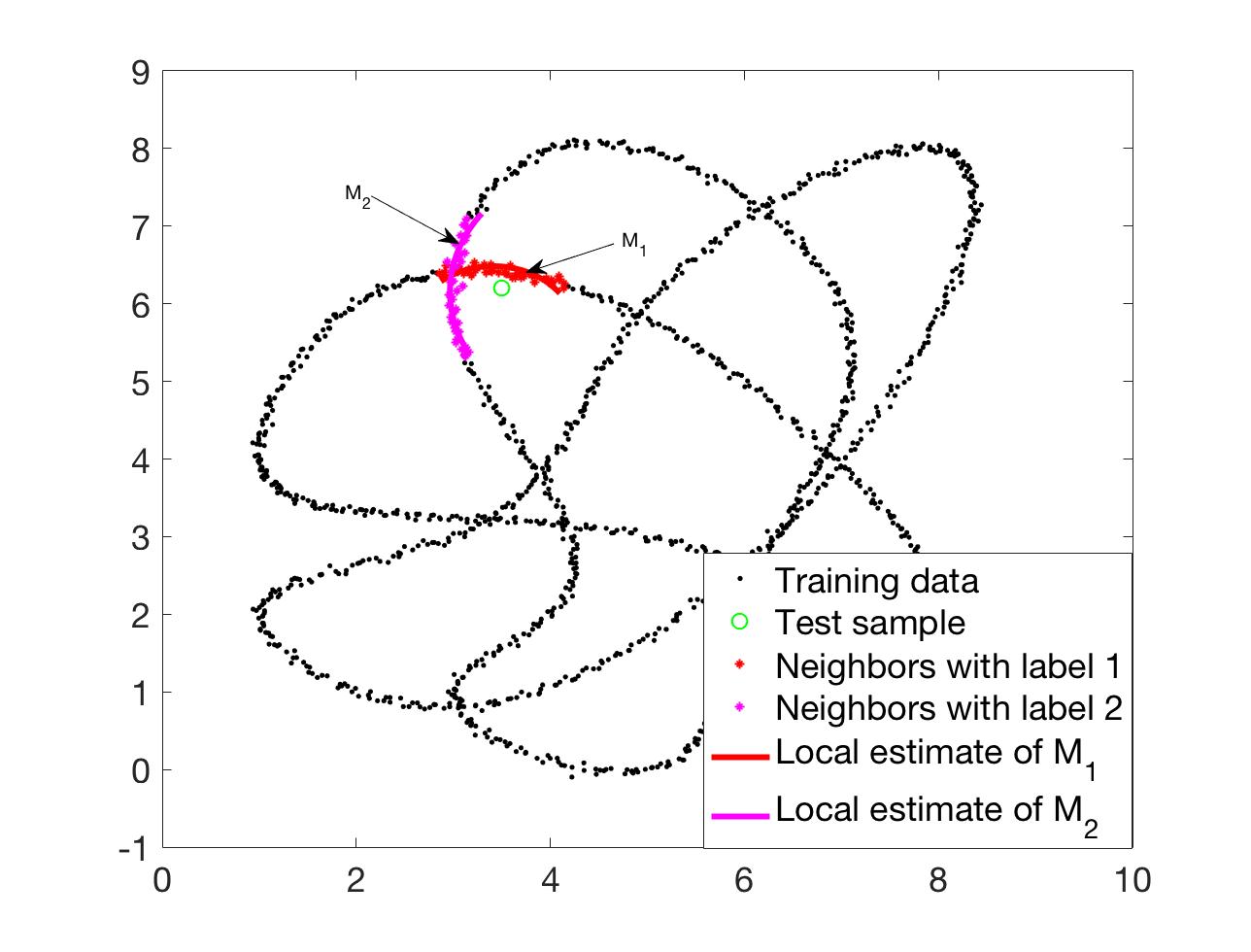} 
			\caption{LOMA classification}
		\end{center}
		\vspace{-0.2cm}
	\end{figure}\label{motiv}

	We use the term `manifold’, since we mathematically assume that the denoised support within a small neighborhood of $x$ is a compact Riemmanian manifold.  This provides a convenient and flexible mathematical representation, which is useful for formalizing the performance of our proposed method.  There is a very rich literature on {\em manifold learning} algorithms ranging from Local Linear Embeddings (LLE) (\cite{lle2000}) to Diffusion Maps (\cite{DM2006}).  The focus of this literature is on non-linear dimensionality reduction, replacing a relatively high-dimensional feature vector $x_i$ with a lower-dimensional set of coordinates $\eta_i$ while maintaining distances between pairs of points.  Manifold learning can be viewed as a non-linear alternative to PCA, and can potentially be used as a first stage dimensionality reduction before applying a classification algorithm.  However, our LOMA approach is fundamentally different from such two-stage approaches, and has much better performance in cases we have considered. One reason could be that we avoid any {\em global} manifold assumption, which is likely much too restrictive in many applications.  
	
	In practice, LOMA requires a specific choice of local manifold approximation.  Motivated by settings with limited training data and by a desire for computational efficiency and transparency, we propose to use spheres for this purpose - obtaining a SPherical Approximation (SPA) classifier.  SPA relies on the Spherical Principal Components Analysis (SPCA) algorithm developed recently by \cite{spherelets}, as a key component of their Spherelets manifold learning algorithm.  Other than using SPCA, the current paper has no overlap with \cite{spherelets}, and the LOMA framework is completely novel to our knowledge.
	
	In Section 2 we provide precise details of the LOMA framework, including the SPA special case, and give theoretical support in terms of asymptotic optimality theory. In Section 3 we apply SPA to the motivating application and two real datasets, and compare SPA to a rich variety of alternative classifiers, showing competitive performance. Proofs are included in the Appendix. Code for implementation of SPA is available at \url{https://github.com/david-dunson/SPAclassifier}.
	
	\section{Local manifold approximation classifier}
	
	Without loss of generality, we focus on the binary classification case for ease in exposition. Assume there are two groups of data, labeled by $1$ and $2$. The features $x_i$ for example $i$ tend to be close to one manifold $M_1$, while features for group $2$ tend to be close to a different manifold $M_2$ and both manifolds are embedded in $\RR^D$ with intrinsic dimension $p<D$. For theoretical purpose in evaluating our proposed method, we assume $x_i=z_i+\epsilon_i$, where $\epsilon_i\sim N(0,\sigma^2 I_D)$ is Gaussian noise and $z_i\in M_{y_i}$ is a denoised location exactly on a manifold. 
	
	Both manifolds may be highly nonlinear, and complex, having varying curvature and even gaps. In addition, the manifolds may be entangled with each other, having multiple intersections and close overlaps; such complexity naturally arises in many real world settings and presents fundamental challenges to current classifiers. 
	
	For a given test sample $x$ to be classified, we calculate the distance between $x$ and the two manifolds, denoted by $d_1$ and $d_2$. Then, we simply assign $x$ to the group with the shorter distance. The key computational step in the LOMA algorithm is the calculation of $d_1$ and $d_2$. In practice, the two manifolds $M_1$ and $M_2$ are unknown, but we have training data containing both $x_i$ and the class label $y_i$, for $n$ different samples. The label $y_i$ is equal to one if example $i$ is in group $1$ and is equal to two if the example is in group $2$. We can use this information to obtain accurate local approximations to the manifolds $M_1$ and $M_2$ within a neighborhood of the feature $x$ to be classified. One can potentially consider a broad variety of local approximations, obtaining different versions of LOMA classification. However, from a practical perspective, it is important to use a local approximation that (a) is parsimonious to make efficient use of limited training data and (b) leads to an analytic form for calculation of $d_1$ and $d_2$. Local linear approximations satisfy (a)-(b) but fail to capture curvature.
	
	\begin{algorithm}
		\SetKwData{Left}{left}\SetKwData{This}{this}\SetKwData{Up}{up}
		\SetKwFunction{Union}{Union}\SetKwFunction{FindCompress}{FindCompress}
		\SetKwInOut{Input}{input}\SetKwInOut{Output}{output}
		\Input{Data $X=\{x_i\}_{i=1}^n$; manifold dimension $p$}
		\Output{Sphere $S_p(V,c,r)$}
		\BlankLine
		\emph{$V = (v_1,\cdots,v_{p+1})$, where $v_j$ is the $j$-th eigenvector of the sample covariance matrix of $X$ when the eigenvalues are in decreasing order}\;
		\emph{$\xi_i=\bar X+VV^{\top}(x_i-\bar X)$ where $\bar X$ is the sample mean of $X$}\;
		\emph{$c=-\frac{1}{2}\left\{\sum_{i=1}^n\left(\xi_i-\bar \xi)(\xi_i-\bar \xi\right)^\top\right\}^{-1}\left\{\sum_{i=1}^n \left(\xi_i^\top  \xi_i-\frac{1}{n}\sum_{j=1}^n\xi_j^\top \xi_j\right)\left(\xi_i-\bar \xi\right)\right\}$}\;
		\emph{$r=\frac{1}{n}\sum_{i=1}^n\|\xi_i-c\|$}\;
		
		\caption{SPCA algorithm}
		\label{SPCA}
	
	\end{algorithm}

	With this motivation, we instead use local spherical approximations. Spheres are simple geometric objects that are easy to fit and work with, providing a generalization of hyperplanes that can dramatically improve accuracy through approximating the local curvature. The center, radius and dimension of each sphere are optimized to provide the best approximation. The distance $d_1$ and $d_2$ can then be easily calculated relying on the spherical approximation. Let $X_{[K]}^l$ be the $K$ nearest neighbors of $x$ among samples with label $l$. We fit a sphere to these points using SPCA, obtaining $\widehat{M}_l$ as the local manifold approximation around $x$ in class $l$. We then approximate $d_l$ by $\widehat{d}_l\coloneqq d(x,\widehat{M}_l)$ and choose the label $y$ as the value of $l$ having the smallest $\widehat{d}_l$. When applied to data $X=\{x_i\}_{i=1}^n$, SPCA produces an estimated $p$-dimensional sphere $S_p(V,c,r)$ having center $c$ and radius $r$ lying in subspace $V$. $V$, $c$ and $r$ are obtained by Algorithm 1. Algorithm 2 provides pseudo code to implement the SPA algorithm.

	\begin{algorithm}
		\SetKwData{Left}{left}\SetKwData{This}{this}\SetKwData{Up}{up}
		\SetKwFunction{Union}{Union}\SetKwFunction{FindCompress}{FindCompress}
		\SetKwInOut{Input}{input}\SetKwInOut{Output}{output}
		\Input{Training data $\{x_i,y_i\}_{i=1}^n$; parameters $K,\ p$; test data $x$ to be classified}
		\Output{Predicted label $y$ for $x$}
		\BlankLine
		Let $L$ be the number of classes\;
		\For{$l=1:L$}{
			\emph{Find $K$ nearest neighbors of $x$ with label $l$, denoted by $X^l_{[K]}\subset \{x_i|y_i=l\}$}\;
			\emph{Calculate the $p$-dimensional spherical approximation of $X^l_{[K]}$, denoted by $S^l_p(V_l,c_l,r_l)$} given by Algorithm \ref{SPCA}\;
			\emph{Calculate the projection of $x$ to the sphere $S^l_p(V_l,c_l,r_l)$, $\widehat{x}^l=\mathrm{Proj}^l(x) = c_l+\frac{r_l}{\|V_lV_l^\top(x-c_l)\|}V_lV_l^\top(x-c_l)$}\;
			\emph{Calculate the distance between $x$ and the sphere $S^l_p(V_l,c_l,r_l)$, $\widehat{d}_l = \|x-\widehat{x}^l\|$}\;
			
		}
		\emph{Assign $x$ to the group with the smallest distance: $y=\underset{l=1,\cdots,L}{argmin}\ \widehat{d}_l$}.
		\caption{SPA algorithm}
		\label{SPAalg}

	\end{algorithm}
	SPA is designed to be simple and directly targeted to detect differences in non-linear support across groups, leading to substantial gains when training data are limited in size. SPA can learn quickly with fewer training data and requires no manual tuning. The only tuning parameters are $K$, the size of the local neighborhoods, and $p$, the dimension of the manifold approximating the denoised support of the data. $K$ can be set to be a default value dependent on $n$ to avoid tuning, while $p$ is easy to tune automatically in being a small integer. This tuning can be done on a subset of the training data, adding a negligible computational cost. Also, SPA has theoretical guarantees on classification performance as training sample size $n$ grows, given by the following two theorems correspond to clean data and noisy data. 

	\begin{theorem}
	Let $M_1$ and $M_2$ correspond to two compact Riemannian manifolds.  Assume $\{x_i\}_{i=1}^n\stackrel{iid}{\sim} \rho$ and $\operatorname{supp}(\rho)=M_1\cup M_2$, $x_i\in M_{y_i}$. Given a test sample $x$ with true label $y$, let $\widehat{y}_n$ be the predicted label obtained by the SPA classifier, then
	\begin{equation*}
		\lim_{n\to\infty}\mathbb{P}(y\neq\widehat{y}_n)\leq\rho(M_1\cap M_2).
	\end{equation*}
	\end{theorem}

In Theorem 1, the data in class $j$ are assumed to take values in $M_j$, which is a compact Riemannian manifold without noise. The overall density of the data across the classes is $\rho$. The theorem shows that as the training sample size $n$ increases, the probability the algorithm produces exactly the correct class label is eventually greater than $1-\rho(M_1\cap M_2)$, where $\rho(M_1\cap M_2)$ is the probability $\rho$ assigns to the intersection region between $M_1$ and $M_2$.

As a corollary, when $\rho(M_1\cap M_2)=0$, the limit is one. This means that SPA will have perfect classification performance for sufficiently large training sample sizes as long as the classes are geometrically separable or the intersection region has measure zero. Theorem $2$ considers the noisy case, which is more realistic in most applications.

\begin{theorem}
Let $M_1$ and $M_2$ correspond to two compact Riemannian manifolds.  Assume $\{z_i\}_{i=1}^n\stackrel{iid}{\sim} \rho$ and $\operatorname{supp}(\rho)=M_1\cup M_2$, $z_i\in M_{y_i}$,  $x_i=z_i+\epsilon_i$, where $\epsilon_i\sim N(0,\sigma^2 I_D)$. Given a test sample $x$ with true label $y$, let $\widehat{y}_n$ be the predicted label obtained by the SPA classifier. Let $\delta>0$ and $B_\delta(M)\coloneqq\{x|d(x,M)<\delta\}$, then
$$\lim_{n\to\infty}\mathbb{P}(y\neq\widehat{y}_n)\leq\rho(B_\delta(M_1)\cap B_\delta(M_2))+\exp\left\{-\frac{\delta^2}{8\sigma^2}+\frac{D}{2}\log\left(\frac{\delta^2}{4\sigma^2}\right)-\frac{D}{2}\left(\log(D)-1\right)\right\}.$$

\end{theorem}

In Theorem 2, the data in class $l$ are distributed around $M_l$ with Gaussian noise.  The theorem shows that as the training sample size $n$ increases, the probability the algorithm produces the wrong class label is asymptotically bounded by
$$
 \rho(B_\delta(M_1)\cap B_\delta(M_2))+\exp\left\{-\frac{\delta^2}{8\sigma^2}+\frac{D}{2}\log\left(\frac{\delta^2}{4\sigma^2}\right)-\frac{D}{2}\left(\log(D)-1\right)\right\}.$$
As the noise level decays to zero, $\sigma\to 0$, let $\delta=\sqrt{\sigma}\to 0$ so the first term converges to $\rho\left(M_1\cap M_2 \right)$. Since $\frac{\delta^2}{\sigma^2}\to\infty$, the second term converges to $0$, so the bound in Theorem 2 coincides with the bound in Theorem 1. This is not surprising since Theorem 1 is a special case of Theorem 2, that is, the case when the noise is zero. The quantity $\frac{\delta^2}{\sigma^2}$ can be viewed as the ``signal-to-noise'' ratio in this setting. The larger $\frac{\delta^2}{\sigma^2}$, the better the performance of the SPA classifier.

	\section{Numerical Examples}
	\subsection*{Funky Curves}
	We first consider the example from Figure 1b. For $y_i=l$, we let $x_i=z_i+\epsilon_i$, where $z_i\in M_l$ is a randomly generated point on a highly non-linear curve $M_l$ and $\epsilon_i$ is a zero mean Gaussian noise. Half of the data are reserved as a test set and a proportion of the other half is used in training. The curves $M_1,M_2,M_3$ are entangled and overlapping, making the classification problem highly challenging. Most classification algorithms are completely unable to deal with entangled, overlapping and intersecting non-linear supports. SPA is specifically designed to easily accommodate this problem, and hence should outperform other algorithms in this and related settings. Complex black box algorithms, such as deep neural networks, can eventually do well, but require much larger training sample sizes. Figure 3 shows the accuracy versus training data sample size plot of several competing algorithms including Complex Trees (\cite{quinlan1986dt}), Fine KNN (\cite{cover1967knn}), Fine Gaussian SVM (\cite{crammer2001kernelsvm}) and Deep Neural Networks (\cite{schmidhuber2015deep}).  These were chosen as the best from among dozens of competitors. The plots show that our SPA algorithm has the highest accuracy, which is over $90\%$ when the sample size is only $150$. In addition, an important point is that SPA can produce excellent performance with very limited training data. This is important because in many applied domains, training data are very expensive and one must make do with small samples. SPA thrives in such settings, beating competing classifiers when training data are limited. As sample size increases, carefully tuned Deep Neural Networks will slowly close the gap. Unfortunately, outside of certain specialized settings, labeled training data are a very limited and valuable resource.

	\begin{figure}
		\begin{center}
			\includegraphics[height =135pt, width=250 pt]{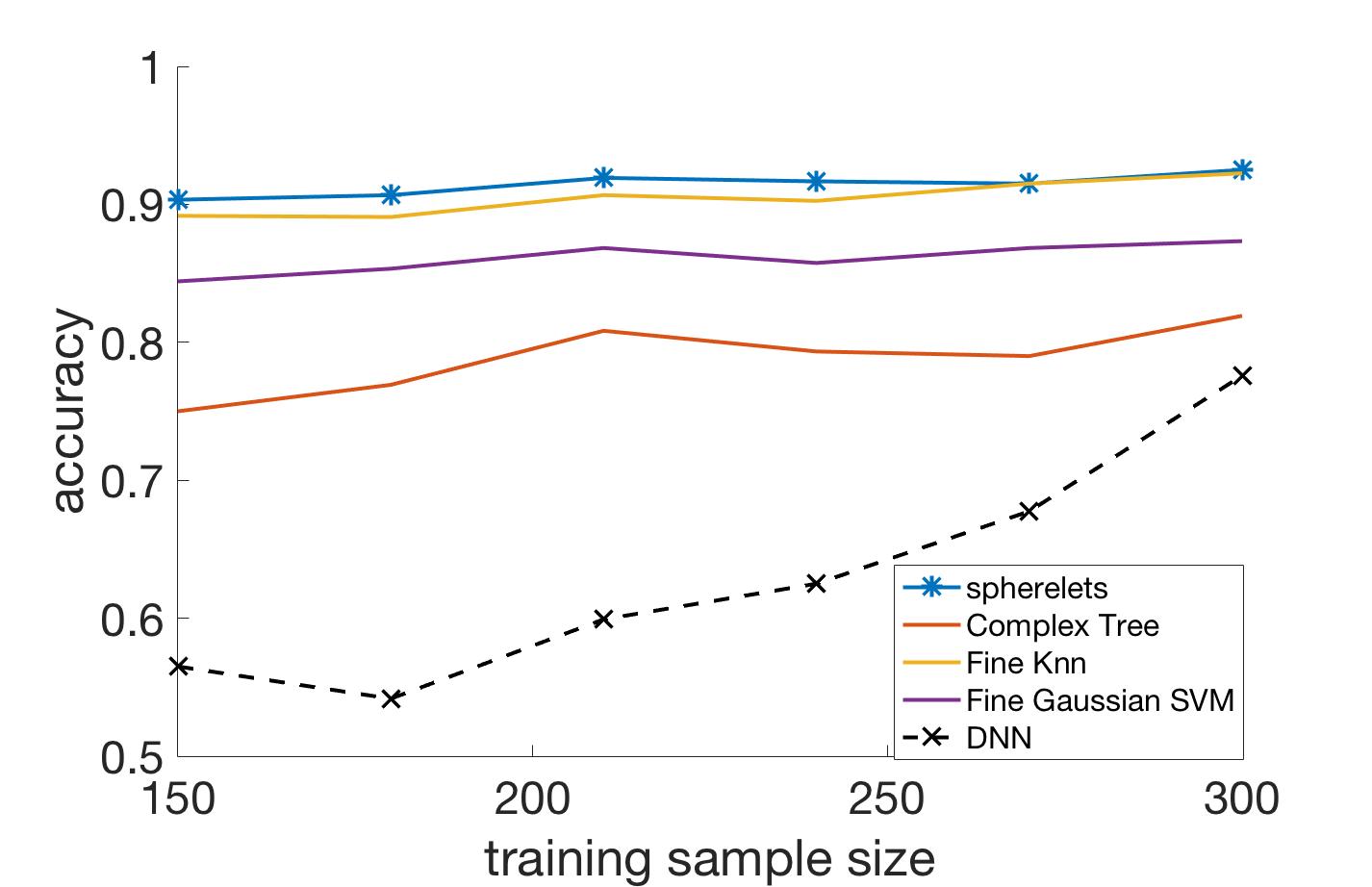} 
			\caption{Accuracy of classifiers on Funky Curves vs training sample size}
		\end{center}
	\end{figure}

	\subsection*{USPS Digits}
	\begin{figure}
		\begin{center}
			\includegraphics[height =135pt, width=400 pt]{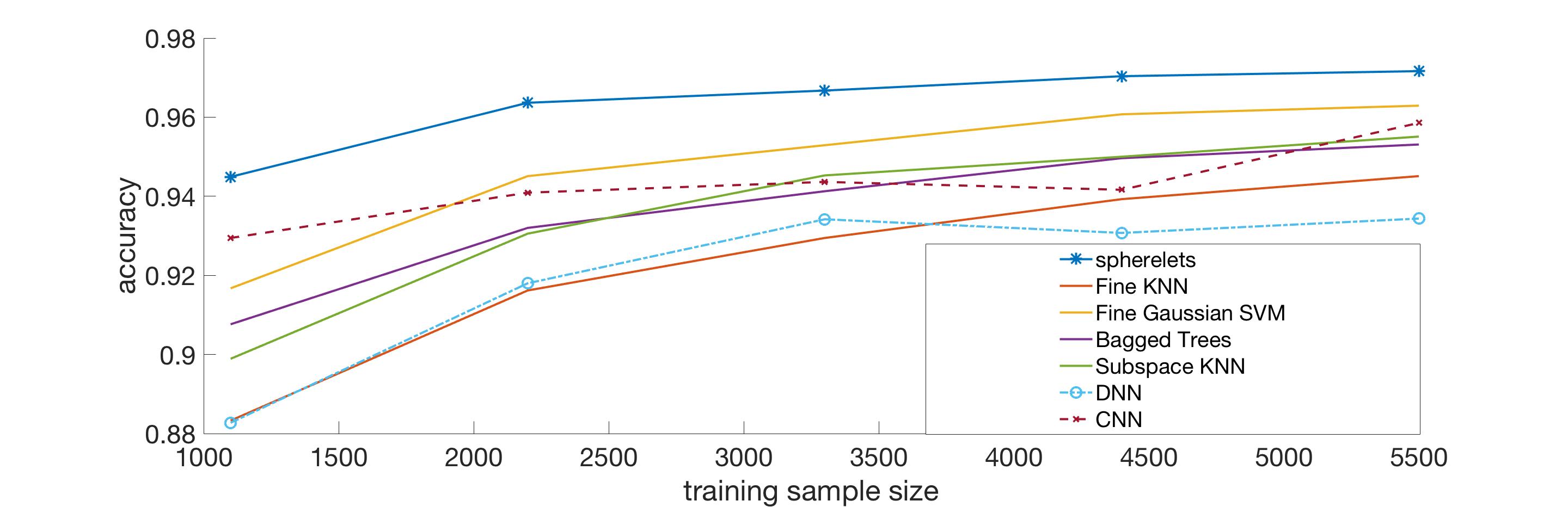} 
			\caption{Accuracy of classifiers on USPS hand written digits vs training data size}	
		\end{center}
	\end{figure}
	We also test our SPA classifier on the USPS digits dataset, one of the standard datasets used for evaluating image classification algorithms. Each sample is an image with $16*16=256$ pixels in gray scale $[0,255]$. There are $11000$ samples and $10$ classes represents digits $0,1,\cdots,9$. When the training sample size is only $10\%$  of the entire set, which is $1100$, the average sample size within each class is only $110$ while the dimension of each sample is $256>110$. Clearly this small sample size is far from enough for complicated algorithms, for example Convolutional Neural Networks (CNN, \cite{ciresan2011cnn}), the most popular algorithm in image classification. Figure 4 shows the accuracy plot, which matches our expectation. When training sample size is only $1100$, SPA has accuracy about $95\%$, higher than competitors.

	SPA works well for this example because although the images have large dimension (256), they are actually lying close to a lower dimension subspace. SPA discovers this lower dimensional geometric structure and simplifies the problem. This ability to learn simple structure in outwardly complex data is a major advantage over black box classifiers, such as DNNs and random forests.

\subsection*{Libras Movement}
\begin{figure}
	\begin{center}
		\includegraphics[height =180pt, width=300 pt]{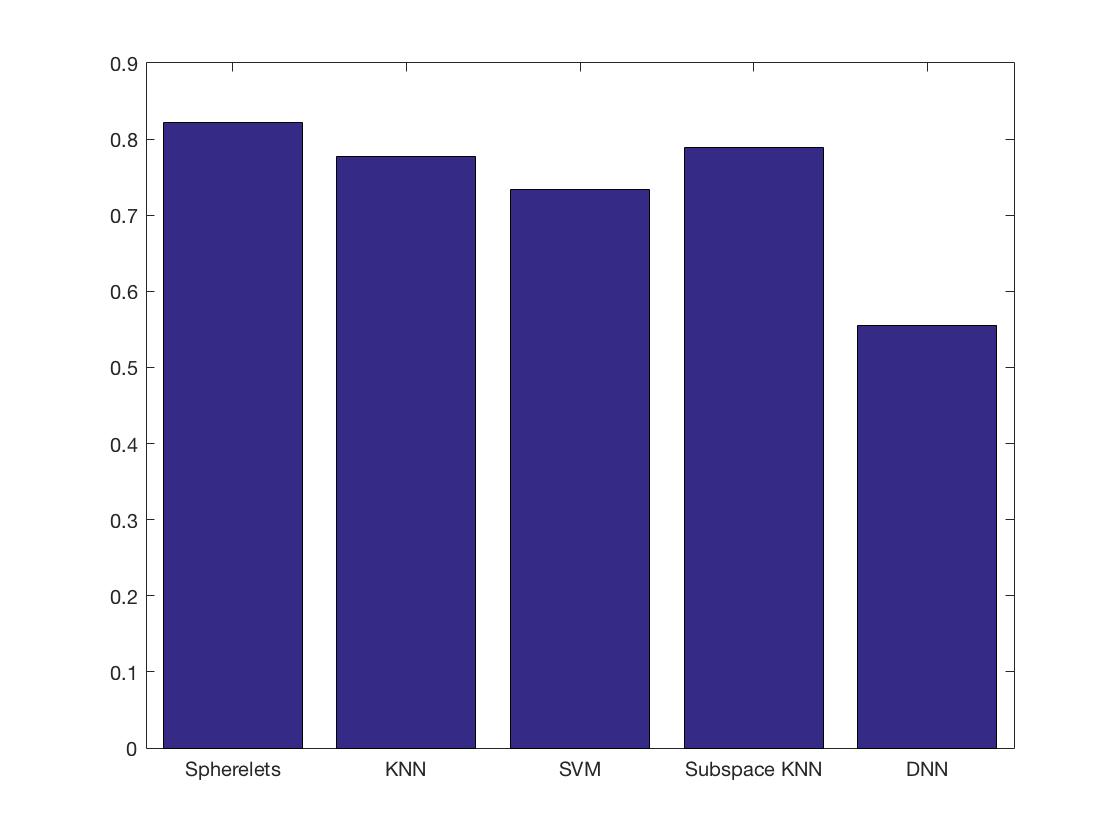} 
		\caption{Accuracy of classifiers on Libras Movement data}	
		
	\end{center}
\end{figure}

The last dataset we consider is the Libras Movement dataset that is available in the UCI Machine Learning Repository. The entire dataset has $15$ classes with $24$ instances for each class, so the total sample size is $360$. Each class represents a type of hand movement in LIBRAS, the official Brazilian signal language. At each time the $2$-d coordinates of the centroid of the hand are recorded and there are $45$ such records so each movement is represented by a $45*2=90$ dimensional vector. As previously, we preserve half the data for testing and use the other half as training samples. We are not varying the training sample size since there are too few samples in this example. For instance, if we use $10\%$ samples to train our model, there will be only $2$ or $3$ samples within each class while the dimension of each sample is $90$. Figure 5 shows that SPA is the only algorithm with accuracy greater than $80\%$ among many other popular algorithms including KNN, Kernel SVM, Subpace KNN and Deep Neural Networks. In this setting, the training sample size is only $180$, that is, there are only $12$ samples within each class while the dimension of the ambient space is $90\gg12$, making this problem even more challenging that the USPS dataset, where the corresponding dimensions are $256\gg110$. The intrinsic dimension $p$ in the SPA classifier is chosen to be $1$ in this example, substantially decreasing the complexity and difficulty of this problem. The intuition for $p=1$ is that since each class represents a certain type of hand movement, which is a $1$-dimensional curve, it is reasonable to assume that $p=1$.

\section{Discussion}
The LOMA framework is quite broad and there are multiple promising directions for future research.  The first is to allow the neighborhood sizes $K$ and/or the manifold dimension $p$ to vary.  This variation can be either local, according to the point $x$ that you want to classify, or global across classes.  Allowing $p_j(x)$ to be varying significantly adds to the flexibility of the approach.  In practice, it is unlikely that most real world datasets have an exact manifold structure even up to iid Gaussian measurement error. Even with a fixed $p$, LOMA is only using the manifold assumption as a local approximation, but still the support may be more complex in certain regions of the feature space than others and for certain classes.  In such cases, there are potential gains to incorporating a varying $p_j(x)$.  However, hurdles include the increased computation and potential need for larger training data for use in tuning $p_j(x)$.  As cross validation is trivially parallelized, the computational barrier is not a major obstacle and one may fit local approximations quickly for different choices of $p_j(x)$ at each $x$ and combine these approximations in an ensemble learning algorithm, avoiding selection of a single manifold dimension.  

An additional direction is to explore alternatives to spheres to develop other special cases of LOMA that may have better performance in certain contexts.  Spheres are remarkably successful at providing a simple modification to hyperplanes that can allow both positive and negative curvature. However, the performance of SPA will decrease when the curvature varies significantly in different directions, for example, image that the point x is on the unit cylinder where the normal curvature ranges from 0 to 1. To accommodate varying curvature, one may use quadratic surfaces as local manifold approximations.  This leads to an increase in the number of parameters needed within each local neighborhood, so that $K$ may need to increase, and it remains to be seen whether efficient computational algorithms can be developed.   

Finally, it is important to generalize LOMA to more complex settings.  Several examples include high-dimensional features in which $D$ is very large, settings in which features are collected dynamically over time, and cases in which features are not real-valued.  To accommodate high-dimensional cases, one straightforward extension is to modify SPCA to use sparse PCA (\cite{zou2006sparse}) within Step 1 of Algorithm 1.  To allow features that are not real numbers, one can instead rely on a distance metric between pairs of features, chosen to be appropriate to the scale of the data, modifying Algorithms 1 and 2 appropriately.

	\subsection*{Acknowledgments} 
	
	The authors acknowledge support for this research from the Office of Naval Research grant N000141712844.
	
	

	\bibliographystyle{apalike}
	\bibliography{ref.bib}

\begin{thebibliography}{}

\bibitem[Ciresan et~al., 2011]{ciresan2011cnn}
Ciresan, D.~C., Meier, U., Masci, J., Maria~Gambardella, L., and Schmidhuber,
  J. (2011).
\newblock Flexible, high performance convolutional neural networks for image
  classification.
\newblock In {\em IJCAI Proceedings-International Joint Conference on
  Artificial Intelligence}, volume~22, page 1237. Barcelona, Spain.

\bibitem[Coifman and Lafon, 2006]{DM2006}
Coifman, R.~R. and Lafon, S. (2006).
\newblock Diffusion maps.
\newblock {\em Applied and Computational Harmonic Analysis}, 21(1):5--30.

\bibitem[Cortes and Vapnik, 1995]{cortes1995support}
Cortes, C. and Vapnik, V. (1995).
\newblock Support-vector networks.
\newblock {\em Machine Learning}, 20(3):273--297.

\bibitem[Cover and Hart, 1967]{cover1967knn}
Cover, T. and Hart, P. (1967).
\newblock Nearest neighbor pattern classification.
\newblock {\em IEEE Transactions on Information Theory}, 13(1):21--27.

\bibitem[Cox, 1958]{logistic}
Cox, D.~R. (1958).
\newblock The regression analysis of binary sequences.
\newblock {\em Journal of the Royal Statistical Society. Series B
  (Methodological)}, pages 215--242.

\bibitem[Crammer and Singer, 2001]{crammer2001kernelsvm}
Crammer, K. and Singer, Y. (2001).
\newblock On the algorithmic implementation of multiclass kernel-based vector
  machines.
\newblock {\em Journal of Machine Learning Research}, 2(Dec):265--292.

\bibitem[Li et~al., 2018]{spherelets}
Li, D., Mukhopadhyay, M., and Dunson, D.~B. (2018).
\newblock Efficient manifold and subspace approximations with spherelets.
\newblock {\em arXiv preprint arXiv:1706.08263}.

\bibitem[Quinlan, 1986]{quinlan1986dt}
Quinlan, J.~R. (1986).
\newblock Induction of decision trees.
\newblock {\em Machine Learning}, 1(1):81--106.

\bibitem[Roweis and Saul, 2000]{lle2000}
Roweis, S.~T. and Saul, L.~K. (2000).
\newblock Nonlinear dimensionality reduction by locally linear embedding.
\newblock {\em Science}, 290(5500):2323--2326.

\bibitem[Schmidhuber, 2015]{schmidhuber2015deep}
Schmidhuber, J. (2015).
\newblock Deep learning in neural networks: An overview.
\newblock {\em Neural Networks}, 61:85--117.

\bibitem[Zou et~al., 2006]{zou2006sparse}
Zou, H., Hastie, T., and Tibshirani, R. (2006).
\newblock Sparse principal component analysis.
\newblock {\em Journal of computational and graphical statistics},
  15(2):265--286.

\end{thebibliography}

	\section*{Appendix}

	\subsection*{Proof of Theorem 1}

	\begin{proof}
		Assume $x\in M_1\setminus M_2$, so $d_1\coloneqq d(x,M_1)=0$ but $d_2\coloneqq d(x,M_2)>0$ by the compactness of $M_2$. Let $\widehat{d}_l$ be the estimate of $d_l$ obtained by the SPA classifier. From Corollary 3 in \cite{spherelets}, we know that $\widehat{d}_l\to d_l$ in probability as $n\to \infty$ for $l=1,2$ so $\lim_{n\to\infty} \mathbb{P}\left(\widehat{d}_2> \widehat{d}_1\right)=1.$
		Recalling the definition of $\widehat{y}_n$, we know that $\widehat{y}_n=y=1 \Longleftrightarrow \widehat{d}_2> \widehat{d}_1$. Hence we conclude that 
		$\displaystyle{\lim_{n\to\infty}\mathbb{P}(y=\hat{y}_n)=\lim_{n\to\infty}\mathbb{P}\left( \widehat{d}_2> \widehat{d}_1 \right)=1}.$
		A similar equation holds for all $x\in M_2\setminus M_1$. Combining the above two situations, we conclude that as long as $x\notin M_1\cap M_2$, the prediction is correct asymptotically, so we have the desired result $$\lim_{n\to\infty}\mathbb{P}(y\neq\hat{y}_n)\leq\rho(M_1\cap M_2).$$
	\end{proof}

	\subsection*{Proof of Theorem 2}

		\begin{proof}
	Without loss of generality, assume $y=1$ so $z\in M_1$. By similar argument as in the proof of Theorem 1, we have
	$$\PP(y\neq \widehat{y}_n)=\PP\left(d(x,\widehat{M_1})>d(x,\widehat{M_2})\right)\to \PP\left(d(x,M_1)>d(x,M_2)\right).$$	
	Then observe that for any $\delta>0$,
	\begin{align*}
	\PP\left(d(x,M_1)>d(x,M_2)\right)&=\PP\left(d(x,M_1)>d(x,M_2)|d(z,M_2)<\delta\right)\PP(d(z,M_2)<\delta)\\
	&+\PP\left(d(x,M_1)>d(x,M_2)|d(z,M_2)\geq\delta\right)\PP(d(z,M_2)\geq\delta)\\
&\leq\PP(d(z,M_2)<\delta)+\PP\left(d(x,M_1)>d(x,M_2)|d(z,M_2)\geq\delta\right)\\
&\leq \rho(B_\delta(M_1)\cap B_\delta(M_2))+\PP\left(d(x,M_1)>d(x,M_2)|d(z,M_2)\geq\delta\right). 
	\end{align*}
Then we consider the second term $\PP\left(d(x,M_1)>d(x,M_2)|d(z,M_2)\geq\delta\right)$. Recall that $x=z+\epsilon$, assume $\|\epsilon\|<\frac{2}{\delta}$ and $d(z,M_2)\geq\delta$, then 
$$d(x,M_1)\leq d(x,z)+d(z,M_1)=\|\epsilon\|<\frac{\delta}{2},$$
while 
$$d(x,M_2)\geq d(z,M_2)-d(x,z)\geq \delta-\|\epsilon\|>\frac{\delta}{2}>d(x,M_1).$$
As a result, $\PP\left(d(x,M_1)>d(x,M_2)|d(z,M_2)\geq\delta\right)\leq\PP\left( \|\epsilon\|\geq \frac{\delta}{2}\right)=\PP\left(\|\frac{\epsilon}{\sigma}\|^2\geq \frac{\delta^2}{4\sigma^2}\right)$. Since $\epsilon\sim N(0,\sigma^2 I_D)$, $\beta\coloneqq \|\frac{\epsilon}{\sigma}\|^2\sim \chi^2(D)$. For convenient, define  $\alpha=\frac{\delta^2}{4\sigma^2}$, then the tail probability $\PP(\beta\geq \alpha)$ can be controlled by Chernoff's inequality:
$$\PP(\beta\geq \alpha)\leq\exp\left\{-t\alpha-\frac{D}{2}\log(1-2t) \right\}, \  \forall t\in(0,\frac{1}{2}).$$
The right hand side is maximized by $t^*=\frac{1}{2}-\frac{D}{2\alpha}$, so we have the final upper bound for the desired tail probability:
$$\PP\left(d(x,M_1)>d(x,M_2)|d(z,M_2)\geq\delta\right)\leq\exp\left\{-\frac{\delta^2}{8\sigma^2}+\frac{D}{2}\log\left(\frac{\delta^2}{4\sigma^2}\right)-\frac{D}{2}\left(\log(D)-1\right)\right\}.$$

\end{proof}

\end{document}